\documentclass[11pt]{article}

\usepackage[preprint]{acl}

\usepackage{times}
\usepackage{latexsym}
\usepackage{algorithm}
\usepackage{caption}
\usepackage{subcaption}
\usepackage{graphicx}
\usepackage{float}
\usepackage[export]{adjustbox}
\usepackage{multirow}
\usepackage{booktabs}
\usepackage{algpseudocode}

\usepackage[T1]{fontenc}

\usepackage[utf8]{inputenc}

\usepackage{microtype}

\usepackage{inconsolata}

\title{ZzzGPT: An Interactive GPT Approach to Enhance Sleep Quality}

\author{Yonchanok Khaokaew$^1$          \\
  \texttt{y.khaokaew} \\
  \texttt{@unsw.edu.au} \\\And
   Kaixin Ji$^2$\\
   \texttt{ 
      kaixin.ji}\\
      \texttt{@student.rmit.edu.au}\\\And
  Thuc Hanh Nguyen$^1$\\
  \texttt{hanh.t.nguyen}\\
  \texttt{@student.unsw.edu.au}
  \\
  \AND
    Hiruni Kegalle, Marwah Alaofi$^2$\\
    \texttt{hiruni.kegalle@student.rmit.edu.au}\\
    \texttt{marwah.alaofi@student.rmit.edu.au}
   \\\And
    Hao Xue, Flora D. Salim$^1$\\
    \texttt{hao.xue1@unsw.edu.au}\\
  \texttt{flora.salim@unsw.edu.au}\AND
$^1$University of New South Wales,
  New South Wales, Australia\\
   {\bf $^2$RMIT University,  Melbourne, Australia}
}

\begin{document}
\maketitle
\begin{abstract}
This paper explores the intersection of technology and sleep pattern comprehension, presenting a cutting-edge two-stage framework that harnesses the power of Large Language Models (LLMs). The primary objective is to deliver precise sleep predictions paired with actionable feedback, addressing the limitations of existing solutions.
This innovative approach involves leveraging the GLOBEM dataset alongside synthetic data generated by LLMs. The results highlight significant improvements, underlining the efficacy of merging advanced machine-learning techniques with a user-centric design ethos. Through this exploration, we bridge the gap between technological sophistication and user-friendly design, ensuring that our framework yields accurate predictions and translates them into actionable insights. 
\end{abstract}

\section{Introduction}

Sleep, an often underestimated factor, plays a pivotal role in this holistic understanding of health. Its quality has profound effects on mental sharpness, emotional balance, productivity, and overall life satisfaction. The advent of wearable sensor technologies has been transformative, offering real-time insights into the complex patterns of sleep and opening new avenues for health improvement.

\begin{figure}[!htb]
    \centering
    \includegraphics[width=0.6\linewidth, trim={3cm 3cm 3cm 3cm}, clip]{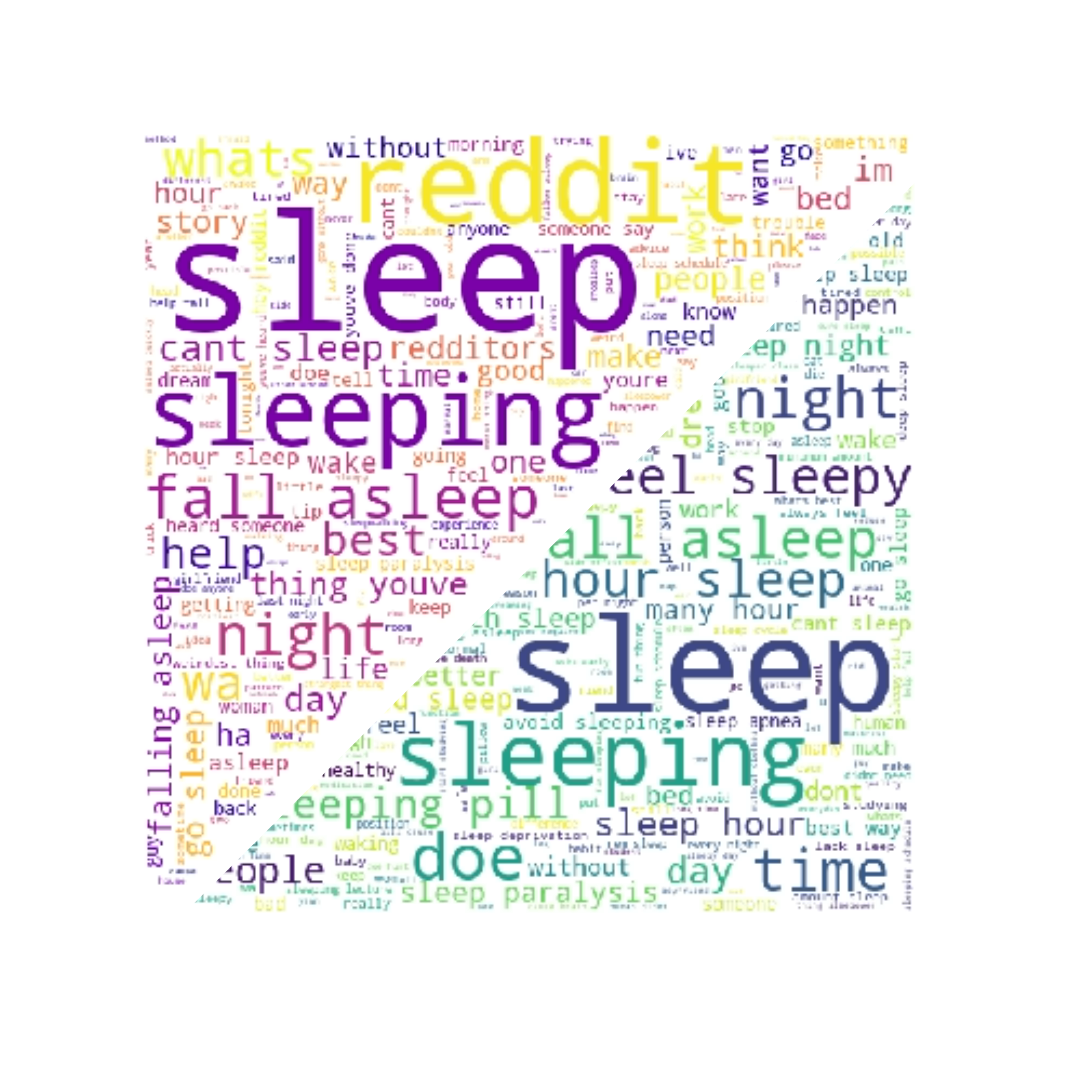}
     \caption{Wordcloud of Quoras~\cite{quora-question-pairs} and Reddit~\cite{redditqa} datasets}
     \label{fig:sleepword}
\end{figure}

Despite these advancements, wearable technologies face significant challenges. They provide a wealth of data yet often fall short in delivering actionable insights, which can lead to user disengagement and device abandonment \cite{attig2020abandonment}. From the user side, the common sleep-related queries on popular online platforms like Reddit \cite{redditqa} and Quora \cite{quora-question-pairs} are usually abstract (See Figure~\ref{fig:sleepword}), such as ``How do I fix/change/maintain a sleep pattern?" and ``I need help to get me to sleep; I start University soon and can't maintain a steady sleep pattern!". This type of question does not clearly describe users' intent or purpose, making it difficult to provide a specific and helpful response. On the other side, such abstract questions also imply users lacking related knowledge \cite{liu2015bed, liang2016qualitative, knowles2018uncertainty} to provide clarification questions \cite{goertzel2014artificial}.

From the system side, potentials of the data to enhance lifestyle and address health issues, such as the intricate relationship between sleep and depression, remains largely untapped. Moreover, a general lack of understanding about the mechanisms of sleep, as well as the algorithms processing the sleep data, further complicates the path to effective health tracking \cite{liu2015bed, knowles2018uncertainty, choe2015sleeptight}.

\begin{figure}[!htb]
     \centering
     \includegraphics[width=\linewidth]{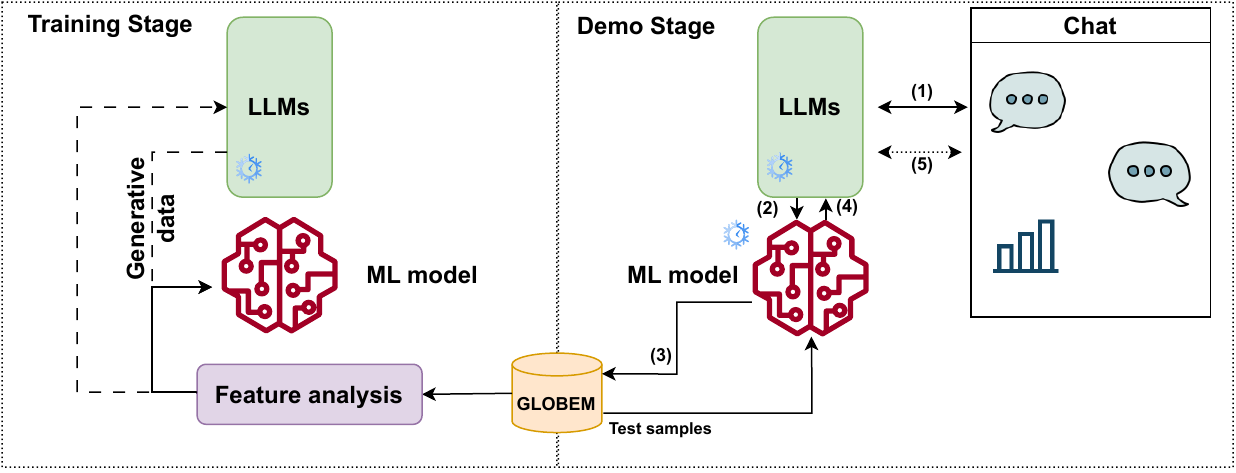}
     \caption{Overview of the proposed two-stage framework}
     \label{fig:framework}
\end{figure}

Addressing these gaps requires not just technological innovation but also a paradigm shift in how we approach and interpret sleep data. This is where Large Language Models (LLMs) come into play.
In the healthcare sector, LLMs play a significant role in enhancing the benefits of pervasive computing by providing more accurate and personalized patient care through interpreting patient feedback, clinical notes, and real-time health data \cite{mesko2023imperative}. They are used in various applications ranging from chronic disease management to mental health assessment \cite{eggmann2023implications,abbasian2023conversational,han2023randomized}.
This project aims to present a holistic, interactive model to help users better understand their sleep, by utilizing the power of Large Language Models (LLMs). In considering explainability, we frame the problem as a regression task where sleep efficiency is graded on a scale of 0 to 100. We propose a detailed two-stage framework 
(Figure~\ref{fig:framework}), where LLMs play different roles at each stage:
\begin{description}
    \item Stage One, a \textbf{\textit{Predictive model}} (discussed in Section~\ref{sec:training}) is trained using the GLOBEM dataset, a comprehensive longitudinal dataset, to identify and filter features correlated with sleep quality, as a regression task. We have also identified an issue of data imbalance. incorporating synthetic data from LLMs to boost predictive accuracy. The experiment result is discussed in Section~\ref{sec:experiments}. 
    \item Stage Two, an \textbf{\textit{Interactive Interface}} (discussed in Section~\ref{sec:demo}), a chatbot application, is developed \footnote{\url{https://github.com/cruiseresearchgroup/ZzzGPT-ACL-2024-}}., providing a user interface with functionalities like sleep quality prediction, data acquisition, interactive responses, and personalized recommendations.
\end{description}

\section{Dataset \& Problem Definition}
\label{sec:problem}

We use data and features from the GLOBEM dataset \cite{xu2022globem}, a longitudinal collection of comprehensive mobile and wearable sensing data. The mobile data includes location, screen status, Bluetooth scans, and call logs. The wearable data were collected via Fitbit devices to track physical activity and sleep patterns. 
The data collection ran four times from 2018 to 2021, with participants contributing data for an average of 78 days per year. 


We choose the feature \textit{`avgefficiency'} as our target variable. This feature is a percentage score calculated by Fitbit based on various sleep features, e.g., sleep or awake duration and sleep stages, to quantify sleep efficiency \footnote{\url{https://www.rapids.science/1.6/features/fitbit-sleep-summary/}}. We term this feature as sleep efficiency from now on.


\begin{figure}[h!]

         \centering
         
        
     \includegraphics[width=0.75\linewidth]{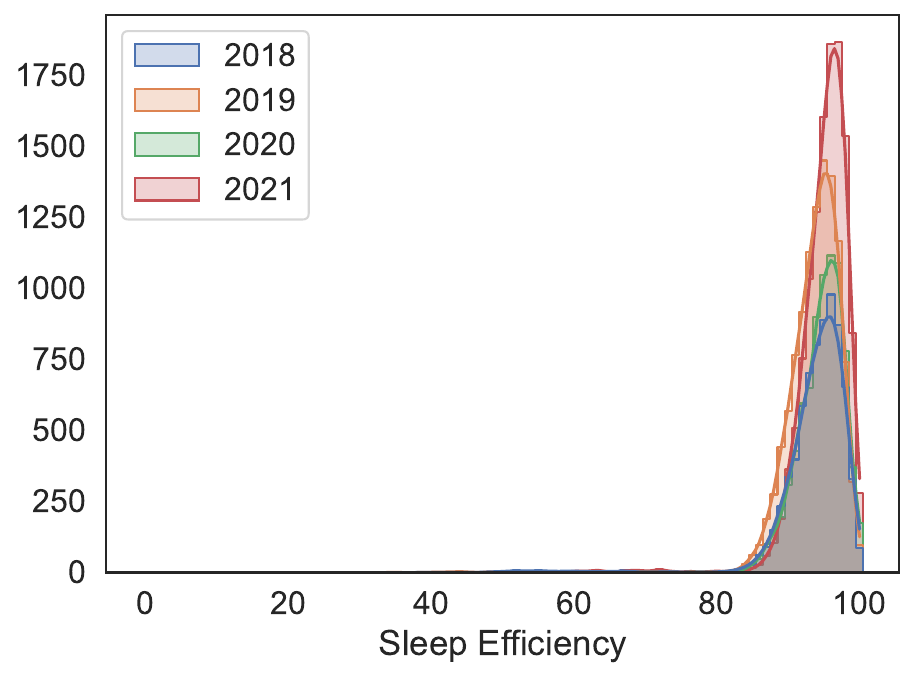}

  \caption{Sleep Efficiency distribution throughout 4 years of GLOBEM dataset (2018 to 2021)}
  \label{fig:sleepeffdis}
\end{figure}

Before building the model, we examine the distribution of sleep efficiency scores. 
As presented in Figure~\ref{fig:sleepeffdis}, the vast majority of scores between 70\% and 90\% indicate fairly good to excellent sleep quality, which obscures those with poor sleep, who need the most intervention. This issue of under-representative might reduce the model's ability in learning sleep patterns, and cause biased predictions. This is also a common issue for real-life data. 
To address this, incorporating LLMs could enhance data representation by generating synthetic examples of lower-efficiency sleep, aiding in developing balanced, fair predictive models.

LLMs can analyze complex data interrelations, uncovering subtle patterns for personalized health insights. This approach not only balances the dataset but also deepens the understanding of individual sleep patterns, supporting tailored health interventions. The aim is to refine predictive models to offer nuanced, actionable advice for all participants, capturing broad trends while focusing on individual sleep efficiency nuances.

\section{Stage One -- Predictive Model}
\label{sec:training}

The model is designed to predict sleep quality $Q_x$ for a user instance $x$ during a time slot $t$, based on a function $f(F_r, F_l) \rightarrow Q_x$, where $F_r$ includes raw features like heart rate and movement, and $F_l$ represents latent factors enhanced by LLMs. The model is formulated as:

\begin{equation}
f(F_r, F_l) \rightarrow (Q_x)
\end{equation}

The use of LLMs to augment the dataset targets the skewness in sleep data, generating synthetic examples of less common sleep patterns to create a balanced dataset for training. This approach not only improves the model's accuracy but also provides a richer analysis of sleep dynamics, leveraging LLMs to uncover complex factors influencing sleep quality, thus fostering a comprehensive understanding and enhancement of sleep health.

A hypothesis is proposed: \textit{Could synthetic data generated from state-of-the-art Large Language Models (LLMs) enhance the model's predictive power?} These LLMs are adept at replicating the natural variance of human behaviour and environmental interactions. Therefore, by introducing this method, we aim to test whether such augmentation would bolster the model's performance or inadvertently introduce noise and complexities.


     


         



\subsection{Regression Models}

Given the nature of our research objective, we approach this task as a regression problem. We evaluate various regression models, including 
Random Forest (RF) \cite{biau2012randomforest}, CatBoost (CatB) \cite{prokhorenkova2018catboost}, and Gradient Boosting (XGB) \cite{chen2016xgboost}. We also try an Ensemble Model composed of the CatB, RF and Ridge Regressor to evaluate and compare their performance. In the Ensemble Model, a stacking regressor is employed to optimize the weighting of predictions from each of these models. This involved using the outputs of the CatBoost Regressor, Ridge Regressor, and Random Forest Regressor as features, which were then input into a Linear Regression model to produce the final prediction.

\subsection{Feature Selection}

Because the dataset offers hundreds of features, we first undertake statistical analysis to identify variables that strongly correlate with sleep quality and further filter features using feature importance ranking with two approaches: manually curated features and the top-\(K\) features. 

As our project aims to provide sleep insights, we carefully select features by considering 1) supported by relevant literature and 2) whether the users can understand and adjust to improve their sleep. 
This approach, which involves understanding the reasoning behind the features and grounding them in prior research, can strengthen the reliability of our sleep predictions and avoid issues such as overfitting.
This approach results in 72 features in total. 
The selected features are listed below:
\begin{itemize}

    \item \textbf{surrounding Bluetooth devices}: the proximity and number of surrounding Bluetooth devices can offer insights into an individual's social interactions and ambient environment, both of which have been linked to sleep quality \cite{xu2012will}.
    \item \textbf{phone call}: Frequency, duration, and timing of phone calls can hint at an individual's social commitments and stress levels, which are known factors affecting sleep \cite{shin2017mobile,tettamanti2020long}.
    \item \textbf{location}: an individual's location, particularly visits to green spaces, might suggest relaxation activities beneficial for sleep. Conversely, nighttime urban activity might disrupt sleep. Green spaces are linked with relaxation and potentially improved sleep quality \cite{feng2020impact,yang2020neighbourhood}.
    \item \textbf{screen usage}: excessive screen time, especially before bedtime, correlates with sleep disturbances due to the blue light emitted from screens \cite{wu2015low,christensen2016direct, shin2017mobile}.
    \item \textbf{steps}: physical activity levels, measured through steps, can provide an understanding of fatigue levels, which directly impact sleep quality \cite{bisson2019walk,stepnowsky2011fatigue}.
\end{itemize}

Then, the top-\(K\) is chosen by comparing the influence on the performance with different \(K\) and the best-perform model. Lastly, the top 20 features (from CatB) perform the best. We have also tried the permutation feature importance and attention-based feature importance from a Long short-term memory (LSTM) model \cite{sathyanarayana2016sleep}; however, the results are not better than the ML model-based approach.
The inclusion of LLM-generated data in the training set is also investigated for its potential performance boost.

\subsection{Data Pre-processing}\label{sec:preprocessing}
Each data entry is mapped with participant ID and recorded date. 
The entries with missing sleep efficiency are excluded. The features with over 50\% missing data are also removed. 
Outliers on some critical features, e.g., screen time and step count, are detected with the Tukey method then imputed with respective participant mean values. 
The features weakly correlated ($r<.001$) with the target variable are discarded. 
Historical data integration requires omitting the initial seven entries per participant. We also add `day of the week' and `weekend' features to capture weekly patterns and tackle multicollinearity by removing one of each feature pair with a correlation above 0.8. 
This stage results in 72 features for subsequent analysis steps.
After applying $z-score$ standardization, the dataset is split into training and testing sets in an 80/20 ratio.

\subsection{Data Generation with LLMs}
Then, we employ the predictive capabilities of LLMs. Our primary goal is to emulate the intricate patterns seen in the actual participant data. Starting with sample selection, a subset comprising 20 samples, equivalent to roughly three weeks of data, is meticulously chosen for each participant. This duration is perceived as adequate for encapsulating an individual's significant behavioural tendencies.

\begin{algorithm}
\caption{Data Generation with LLMs}\label{alg:cap}
\begin{algorithmic}

\Require Samples from participant $S[1..n]$ 
\Ensure Additional dataset list $L[1..N]$

\For{each participant}
\While{counter is less than the number of iterations required}
\State $T$ ← Select 20 samples from participant data (approx. three weeks)
    \State Formulate p1 table-format prompt with these samples $T$;
    \State $cp$ ← $LLM( p1 )$;
    \While{$cp$ does not meet criteria}
    \State $cp$ ← $LLM( p1 )$;
    
    \EndWhile
    \State $L$ ← $L \cup \{F(cp)\}$
    \EndWhile
\EndFor


\end{algorithmic}
\end{algorithm}

\begin{figure}[!htb]
     \centering
     \includegraphics[width=\linewidth]{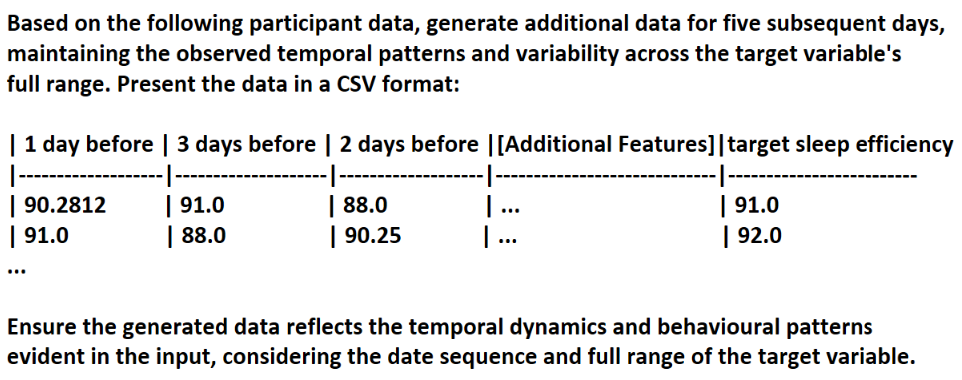}
     \caption{Prompt for generating additional data}
     \label{fig:propmt}
\end{figure}

These chosen samples are methodically transformed into a table-format prompt (see Fig. \ref{fig:propmt}) to guide the LLM's data generation task. This structured presentation offers clarity and ensures the LLM can seamlessly pick up on the dataset's inherent temporal patterns and subtle nuances. With these prompts at the ready, our next course of action involved the OpenAI API, harnessing the power of LLMs to extrapolate from the given data. For each structured prompt based on one participant's data, the model is asked to generate new data for five additional days, taking cues from the preceding week's patterns. However, not all generated data met our rigorous standards. In situations where the generated data did not encompass all requisite features, we manually pruned such data to maintain the dataset's integrity and coherence. The pseudo-code of this stage is outlined in Algorithm \ref{alg:cap}.

Once generated, this new set of data is integrated with the original training set. This augmented compilation provides a larger volume of data and potentially encompasses a more diverse set of behavioural variations. After the augmentation process, we conducted an evaluation to gauge the impact of the LLM-generated data on model performance, keenly observing if it played a role in enhancing the training outcomes. This endeavour aimed to shed light on the potential merits of using LLMs for data augmentation in behavioural predictive modelling.

\section{Experiments}
\label{sec:experiments}

In the following sections, we delve into the experimental framework established to test our hypothesis and evaluate the efficacy of our proposed solution. We outline the specific approach to predicting sleep quality and duration as regression tasks, integrating Large Language Models (LLMs) for data augmentation.

\subsection{Experiment setting}
The primary objective of our study is to forecast the score of \textit{sleep efficiency}. The \textit{sleep efficiency} is the `\textit{average sleep efficiency}', which is recorded by the Fitbit devices, as defined within the GLOBEM dataset. The scope of this experiment is confined to the first cohort year (INS-W\_1), which provides a snapshot of sleep behaviour before the COVID-19 pandemic's onset. Next, we will utilize the procedures described in the Section~\ref{sec:preprocessing}, to preprocess the data, ensuring it's primed for our analytical and predictive modeling tasks.

\paragraph{Baseline Models.}
\label{sec:baseline models}

We establish baseline models to evaluate our method's efficacy in predicting sleep efficiency, drawing on models from existing literature that use the GLOBEM dataset for health-related tasks, especially in depression studies. These models provide a benchmark selected for their relevance to our feature set and their analytical focus on behavioural patterns, which may align with sleep pattern analysis.

\begin{description}
\item [] \textbf{\citet{10.1145/3351274}} The study applies association rule mining on various data points like Bluetooth, calls, and GPS, categorized into time-of-day epochs, and uses an Adaboost Classifier to predict depression, transforming data into categorical variables based on quantile thresholds. Sleep is categorized using a 7-hour threshold to distinguish between different sleep patterns.

\item []\textbf{\citet{7764553}} The study utilizes iOS devices to gather location and activity data to forecast depression, employing a non-linear SVM with an RBF kernel for feature extraction like location variance and entropy. In replication, it incorporates additional data such as step counts and varying settings, utilizing SVM with adjusted parameters for deeper analysis.

\item []\textbf{\citet{info:doi/10.2196/mhealth.5960}} Using a smartphone app, the study collects activity and location data to study depression, using SVM and Random Forest classifiers with two-week historical statistics for analysis. The adaptation focuses on key metrics, excluding text, calendar, and WIFI data, which impacts the sample size due to the exclusion of initial records.

\end{description}

\subsection{Results}

\paragraph{Feature Performance Comparison.}

\begin{table*}[htbp!]
\centering
   \resizebox{0.9\linewidth}{!}{\begin{tabular}{l|ccc|ccc|ccc}
    \toprule

   \textbf{ Features (N)} & \multicolumn{3}{c|} {\textbf{Hand-picked(72)}}&   \multicolumn{3}{c|} {\textbf{Top-k(20)}} &    \multicolumn{3}{c} {\textbf{Top-k(20)+G}}\\
    \midrule
   \textbf{Model/Metric} & $RMSE$ &$MAE$ & $R^2$ &$RMSE$ &$MAE$ & $R^2$ &$RMSE$ &$MAE$ & $R^2$ \\
    \midrule

      Stacked Model &\textbf{ 2.661} & \textbf{2.105} & \textbf{.256}&
                    2.691 & 2.128 &.239 &
                    2.684 & 2.119 &.243 \\ 
    Random Forest & 2.680 & 2.116 & .246 &
                    2.700 & 2.139 & .234 & 
                     2.704 & 2.142 & .232 \\
    XGBoost & 2.704 & 2.138 & .232 & 
                2.704 & 2.138 & .232 &
                2.693 & 2.132 & .238 \\
    Catboost& 2.689 & 2.121 & .239 & 
                \textbf{2.690} & \textbf{2.124} & \textbf{.239} &
                \textbf{2.675} & \textbf{2.114} & \textbf{.248}  \\
  \bottomrule
    \end{tabular}}
   \vspace{1em}
  \caption{Test Performance of models on target Sleep feature. G: LLM-generated training data. Two feature sets are evaluated for each model: \textbf{Hand-picked (72)} and \textbf{Top-k (20)}, and an additional set with new LLM-generated training data \textbf{Top-k (20) + G} (where G denotes LLM-generated training data).}
  \label{tab:testresult_both}
\end{table*}



As presented in Table~\ref{tab:testresult_both}, 
Catboost achieves commendable performance with $RMSE$ of 2.690, $MAE$ of 2.124, and $R^2$ of 0.239. The inclusion of LLM-generated training data (\textbf{Top-k (20) + G}) contributes to improved $R^2$ at 0.248.
Notably, the Stacked Model, Random Forest, and XGBoost, stands out with $RMSE$ of 2.684, $MAE$ of 2.119, and $R^2$ of 0.243 in the \textbf{Top-k (20) + G} scenario.
These results emphasize the impact of different models and feature selection strategies on predicting sleep-related metrics. Moreover, incorporating LLM-generated training data contributes to enhanced predictive accuracy, showcasing the potential of leveraging LLMs in this domain.

\paragraph{Model Performance Comparison.}

 \begin{figure}[th!]
\centering
    \includegraphics[width=0.8\linewidth]{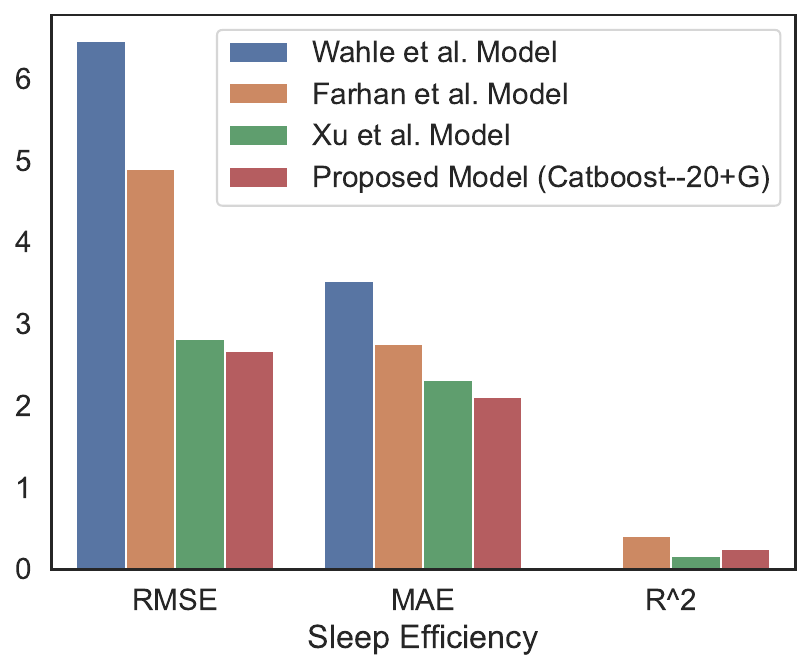}
    \caption{The performance of the proposed model with other baseline models, proposed by \citet{10.1145/3351274}, \citet{7764553}, and \citet{info:doi/10.2196/mhealth.5960}.}
    \label{fig:result_features}
\end{figure}



Figure~\ref{fig:result_features} presents the performance comparison among various sleep prediction models (discussed in Section~\ref{sec:baseline models}).
The \textit{Catboost--20+G} model demonstrates superior performance in predicting sleep efficiency compared to these baseline models, with significantly lower RMSE and MAE values, along with a higher $R^2$ score. This result highlights its accuracy and precision in reflecting the variability of sleep efficiency features. The high $R^2$ score of the \textit{Farhan} baseline model might be due to limited data range diversity, as sleep efficiency is predominantly above 90\%, and there are few low-efficiency samples.

\section{Stage Two -- \textit{Interactive Interface}}
\label{sec:demo}

The Second Stage is the \textit{Interactive Interface} that bridges the users and the predictive model via LLMs. 
Here, LLMs play in 5 roles:

\begin{itemize}

    \item [] \textbf{User Interaction via LLMs}: The Interactive Interface Stage is designed to simulate a real-world application of the trained model. A chat application, powered by a LLM API, acts as the front-end interface. Users see a range of suggested questions when interacting with the chatbot, including queries about their anticipated sleep quality for the night.
    
    \item [] \textbf{Sleep Quality Prediction}: When a user opts to investigate their sleep quality, the system swiftly deploys the predictive model honed during the \textit{Predictive Model} Stage, considering the top 20 important features ranked during analysis to generate a prediction.
    
    \item [] \textbf{Testing Sample Acquisition}: To enact this demonstration, we assume the passive acquisition of relevant user data through the chatbot installed on their devices. A representative testing sample will be culled from the GLOBEM dataset, which comprises the features earmarked as important during the previous \textit{Predictive Model} Stage.
    
    \item []\textbf{Predictive Responses and Interactive Graph}: The system offers an interactive element beyond text-only. The LLM API generates textual advice or comments based on the predicted sleep quality. Furthermore, an interactive graphical interface allows users to modify key features to see their real-time impact on predicted sleep quality.

    \item []\textbf{Personalized Recommendations}: Finally, the application auto-generates pragmatic suggestions tailored to the user's lifestyle and environment. By setting thresholds and manipulating feature values, the system explores how minor adjustments can lead to significant improvements in sleep quality
\end{itemize}

As the demo \footnote{You can find the demo video by clicking on the following URL.: \url{https://www.dropbox.com/scl/fi/s1kftjtfpecwsegbqj16f/ACL-ZZZgpt-demo.mp4?rlkey=mkf5hazotvp0dvhvi1mlwfig6&dl=0}} in Appendix~\ref{sec:appendix} illustrated, upon logging into the application, a user initiates a chat about sleep. However, the question is usually not clear to the chatbot. It seeks more information from the user to clarify their concerns. The chatbot then shares an interactive diagram with the user that shows the features that can influence their sleep. The user can drag the elements to see how they affect their sleep. This will help them identify the factors that are causing their sleep issues. Once the chatbot understands the user's concerns, it provides recommendations to guide them in changing their behaviour, ultimately improving their sleep.

\section{Conclusion}
In conclusion, this project introduces a novel two-stage framework to promote wearable device usage by enhancing explainability and interaction. It combines Machine Learning and large language models (LLMs) with an interactive interface, advancing the use of wearables in health monitoring. Our work contributes to sleep research and personalized healthcare, pioneering predictive modelling. Incorporating data analysis and instant feedback, we've seen encouraging results, highlighting practicality and accessibility.

\section*{ACKNOWLEDGMENTS}
This research is partially supported by the Australian Research Council Centre of Excellence for Automated Decision Making and Society (CE200100005).

\newpage
\section*{Broader impact}
Our proposed framework's envisioned future involves evolving into a personalized health dashboard, akin to established applications such as the Fitbit App~\footnote{https://www.fitbit.com/global/us/technology/fitbit-app}. This transition could mark a paradigm shift in the way individuals engage with their health, offering an intuitive interface that empowers users with actionable insights into their sleep patterns.

Moving forward, the integration of a secure LLM with wearable devices promises a new level of health assistance. With progress in LLM safety, tackling issues like hallucinations \cite{rawte2023survey}, data leaks, and privacy \cite{purtova2015laws}, the envisioned interface could blend smoothly with various wearable devices. This step would mark a new phase in health monitoring, offering users real-time, personalized advice and practical plans to enhance their well-being.

Furthermore, synthesizing sensor data opens avenues for users to engage in meaningful discussions and collaborate with the LLM to develop personalized action plans. This collaborative approach not only enhances the user experience but also empowers individuals to participate in their health journey actively. As we navigate the evolving landscape of health technology, our project stands as a pioneering exploration with far-reaching implications for the future of personalized healthcare and wearable device integration.

\bibliography{main}

\newpage
\appendix

\section{Screenshots of Stage Two Interactive Interface}
\label{sec:appendix}

 \begin{figure}[tbph!]
     \begin{subfigure}[b]{0.45\textwidth}
         \centering
         \includegraphics[width=\textwidth]{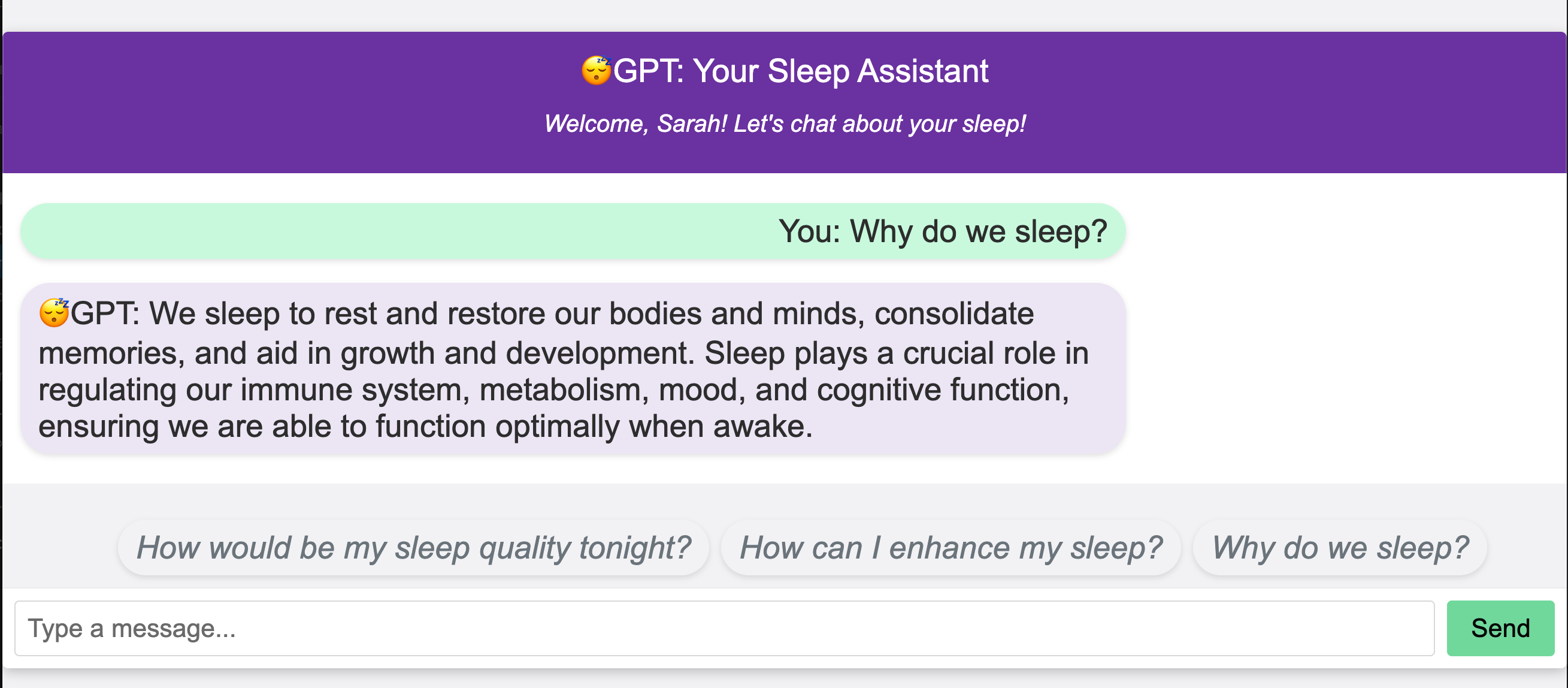}
         \caption{Conversation - 1}
     \end{subfigure}
     \hfill
     \begin{subfigure}[b]{0.45\textwidth}
         \centering
         \includegraphics[width=\textwidth]{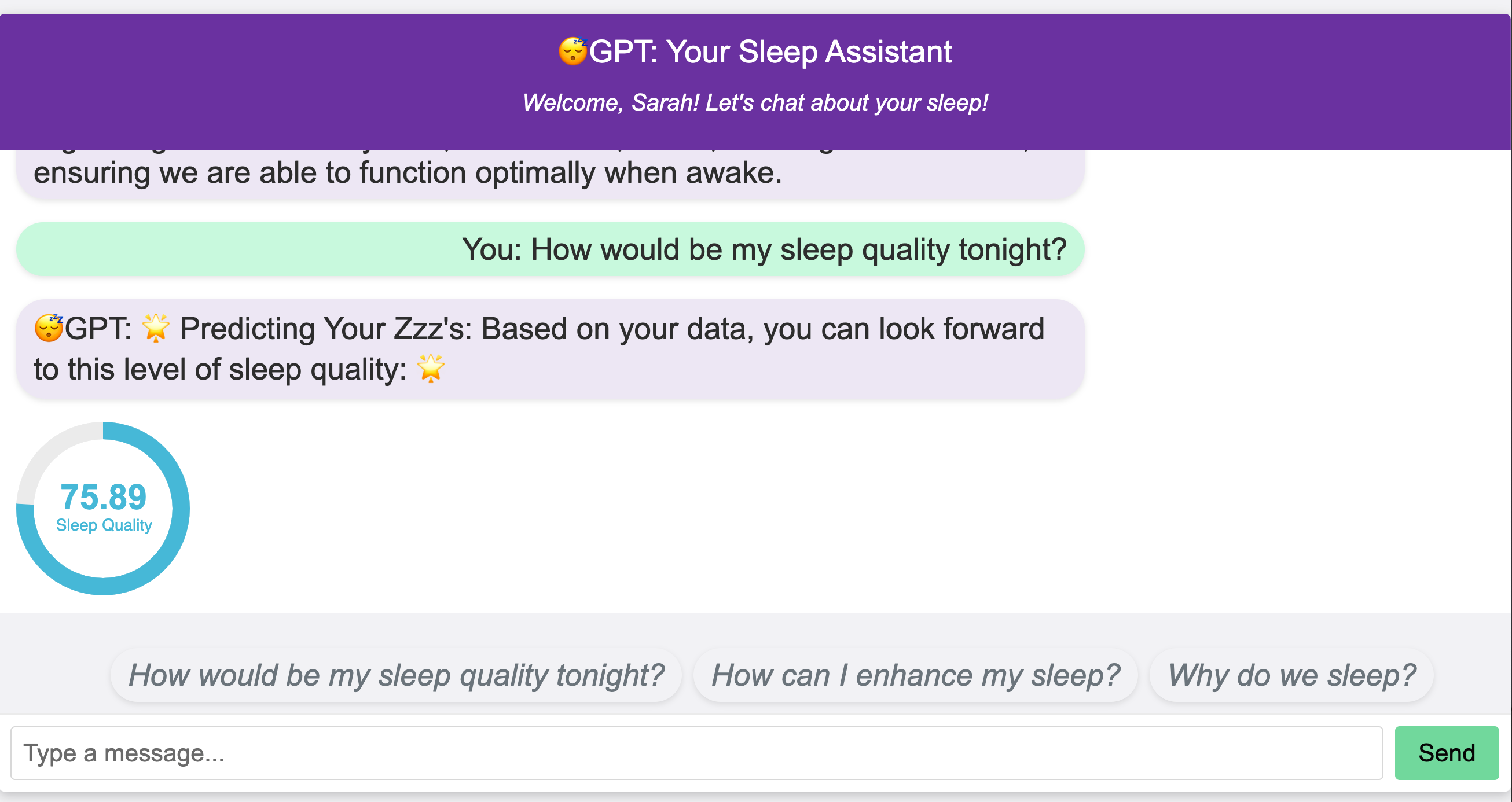}
         \caption{Conversation - 2}
     \end{subfigure}
     \hfill
     \begin{subfigure}[b]{0.45\textwidth}
         \centering
         \includegraphics[width=\textwidth]{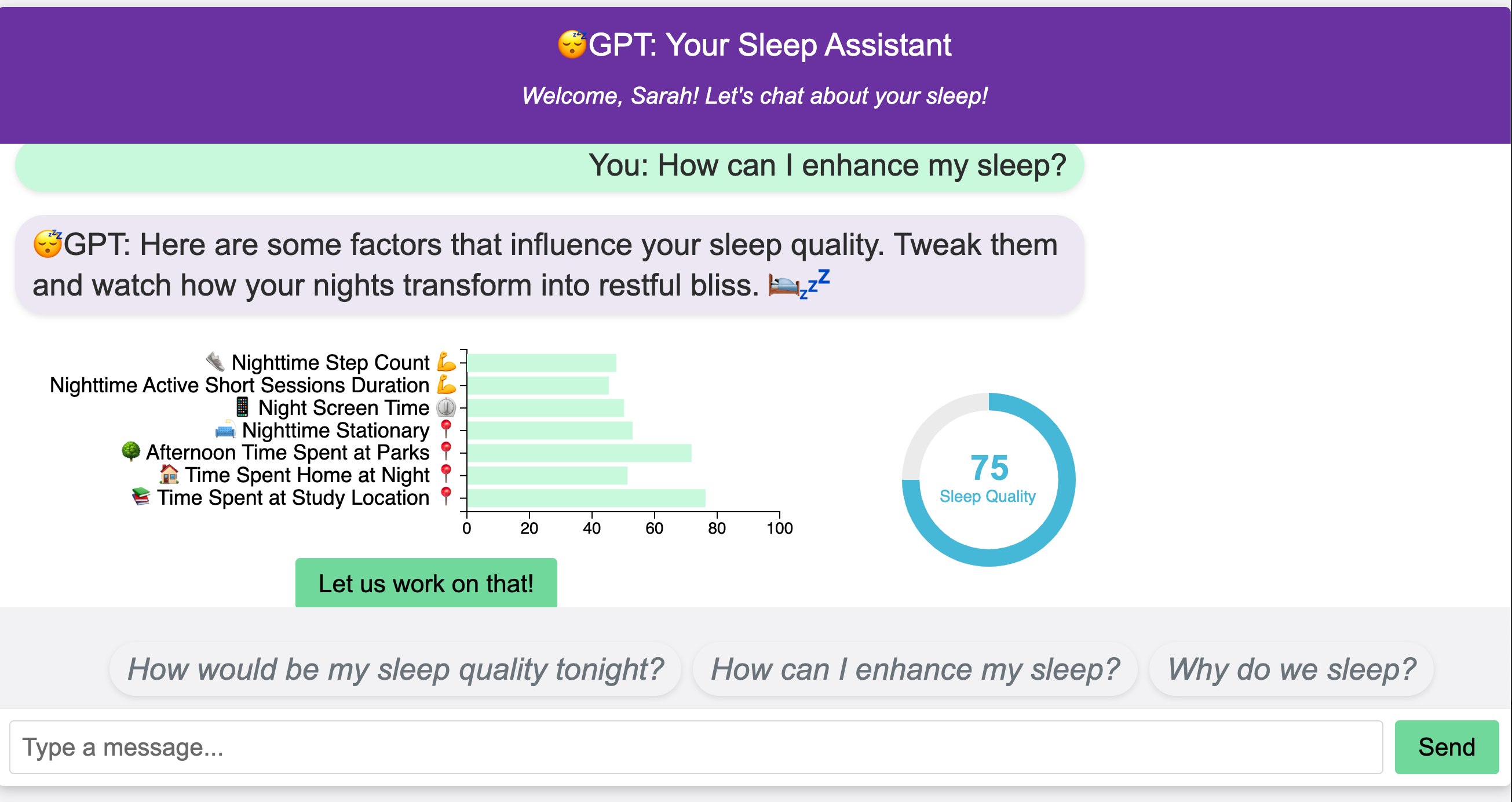}
         \caption{Feature Interaction - 1}
     \end{subfigure}
     \hfill
     \begin{subfigure}[b]{0.45\textwidth}
         \centering
         \includegraphics[width=\textwidth]{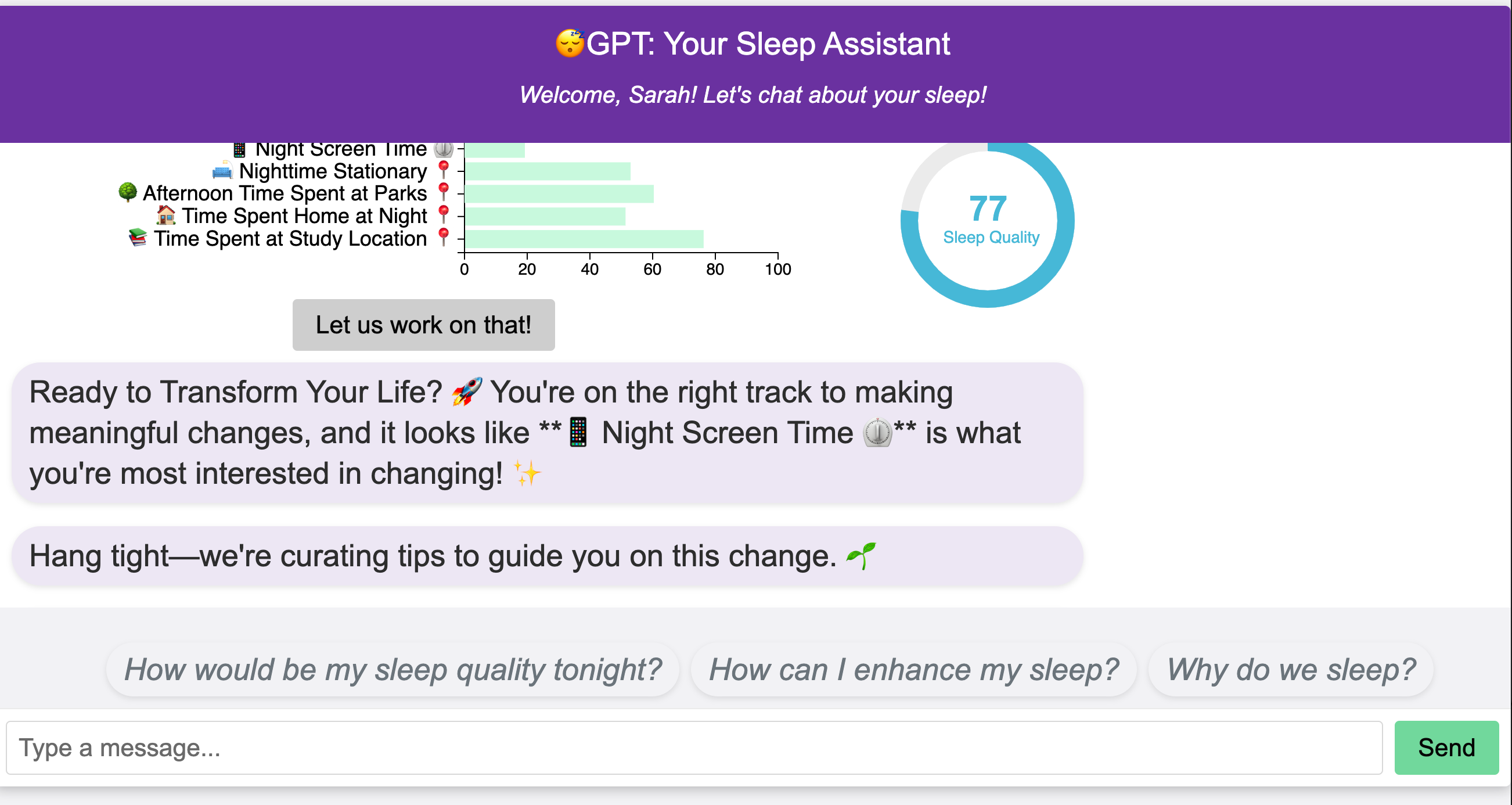}
         \caption{Feature Interaction - 2}
     \end{subfigure}
     \hfill
     \begin{subfigure}[b]{0.45\textwidth}
         \centering
         \includegraphics[width=\textwidth]{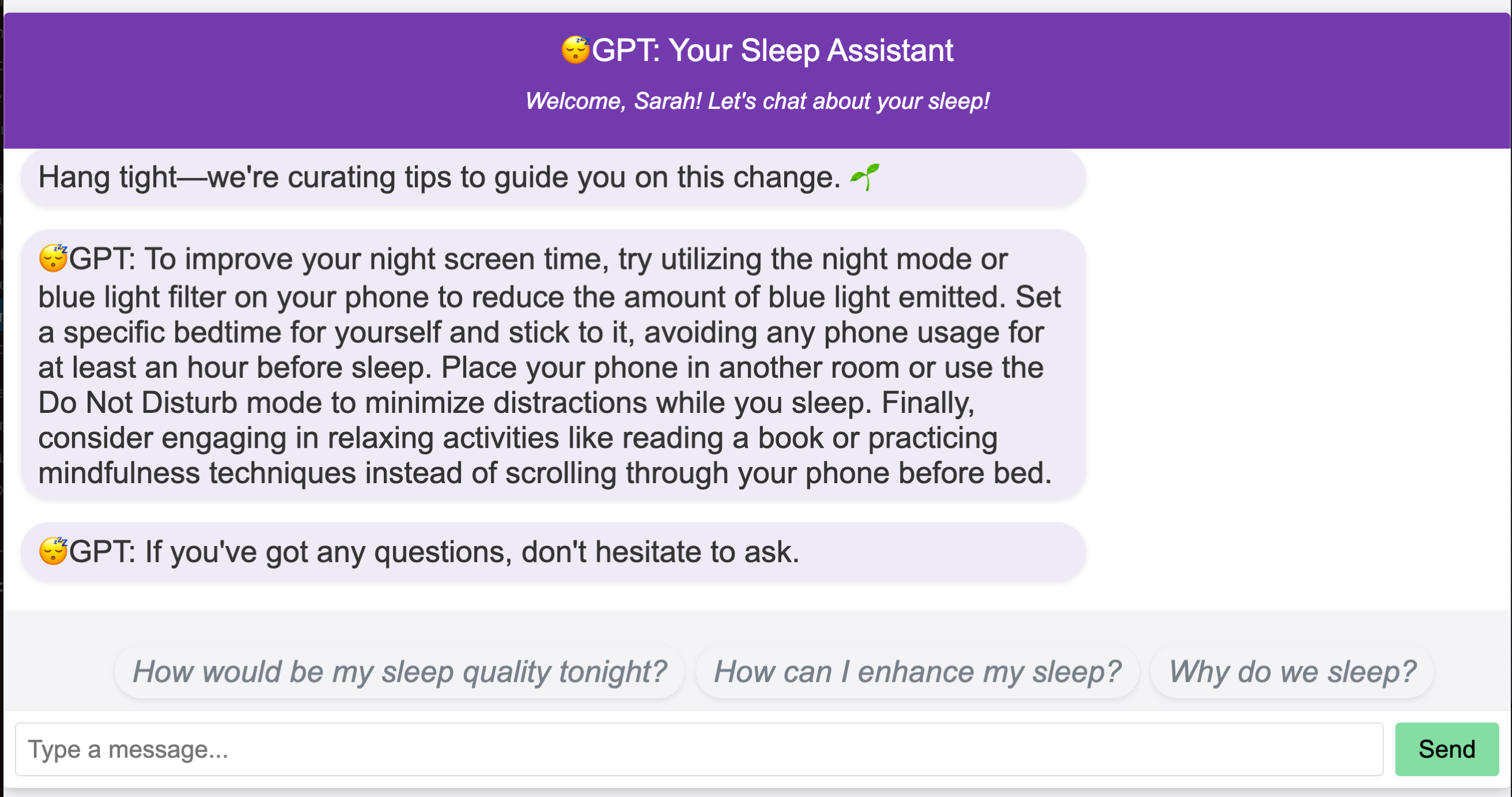}
         \caption{Recommendation}
     \end{subfigure}
     \caption{Sample interaction with the Demo \textit{Interactive Interface}.}
     \label{fig:demo}
 \end{figure}

\end{document}